\pdfoutput=1
\documentclass[letterpaper, 10 pt, conference]{ieeeconf}  
\IEEEoverridecommandlockouts                              
\usepackage{times}
\usepackage[pdftex]{graphicx}
\usepackage{wrapfig}
\usepackage[colorlinks,linkcolor=black,anchorcolor=black,urlcolor=blue]{hyperref}
\usepackage{amsmath,amsfonts,amssymb}
\usepackage{prettyref}
\usepackage{mathrsfs}
\usepackage{graphicx}
\usepackage{wrapfig}
\usepackage{xcolor}
\usepackage{subfig}
\usepackage{MnSymbol}
\usepackage{multirow}
\usepackage{booktabs}
\usepackage{algorithm}
\usepackage{algpseudocode}
\usepackage{slashbox}

\algnewcommand{\LineComment}[1]{\State \(\triangleright\) #1}
\algdef{SE}[DOWHILE]{Do}{doWhile}{\algorithmicdo}[1]{\algorithmicwhile\ #1}
\newcommand*{\colorboxed}{}
\def\colorboxed#1#{%
  \colorboxedAux{#1}%
}
\newcommand*{\colorboxedAux}[3]{%
  \begingroup
    \colorlet{cb@saved}{.}%
    \color#1{#2}%
    \boxed{%
      \color{cb@saved}%
      #3%
    }%
  \endgroup
}

\newrefformat{Fig}{Figure~\ref{#1}}
\newrefformat{fig}{Figure~\ref{#1}}
\newrefformat{par}{Section~\ref{#1}}
\newrefformat{appen}{Appendix~\ref{#1}}
\newrefformat{sec}{Section~\ref{#1}}
\newrefformat{sub}{Section~\ref{#1}}
\newrefformat{table}{Table~\ref{#1}}
\newrefformat{alg}{Algorithm~\ref{#1}}
\newrefformat{Alg}{Algorithm~\ref{#1}}
\newrefformat{Def}{Definition~\ref{#1}}
\newrefformat{Thm}{Theorem~\ref{#1}}
\newrefformat{Lem}{Lemma~\ref{#1}}
\newrefformat{step}{Step~\ref{#1}}
\newrefformat{ln}{Line~\ref{#1}}
\newrefformat{eq}{Equation~\ref{#1}}
\newrefformat{pb}{Problem~\ref{#1}}
\newrefformat{it}{Item~\ref{#1}}
\newrefformat{te}{Term~\ref{#1}}
\def\Eqref Eq:#1:{\eqref{eq:#1}}
\newrefformat{Eq}{Equation~\Eqref#1:}

\newcommand{\E}[1]{\mathbf{#1}}
\newcommand{\TE}[1]{\textbf{#1}}

\newcommand{\sumno}{\sum\nolimits}

\newcommand{\FPP}[2]{\frac{\partial{#1}}{\partial{#2}}}

\newcommand{\TWOR}[2]{\left(\setlength{\arraycolsep}{1pt}\begin{array}{cc}{#1}^T & {#2}^T\end{array}\right)^T}

\newcommand{\fmin}[1]{\underset{#1}{\E{min}}}

\newcommand{\argmin}[1]{\underset{#1}{\E{argmin}}}
\newcommand{\argminP}[1]{\E{argmin}}
\newcommand{\argmax}[1]{\underset{#1}{\E{argmax}}}
\newcommand{\argmaxP}[1]{\E{argmax}}

\newcommand{\TWORCell}[2]{\begin{tabular}{@{}c@{}}#1 \\ #2\end{tabular}}

\newcommand{\GRASPIT}{\textit{GraspIt!}}

\usepackage{cancel}
\usepackage{verbatim}
\makeatletter
\IEEEtriggercmd{\reset@font\normalfont\fontsize{7.0pt}{7.2pt}\selectfont}
\makeatother
\IEEEtriggeratref{1}

\makeatletter
\newcommand\fs@ruled@notop{\def\@fs@cfont{\bfseries}\let\@fs@capt\floatc@ruled
  \def\@fs@pre{}%
  \def\@fs@post{\kern2pt\hrule\relax}%
  \def\@fs@mid{\kern2pt\hrule\kern2pt}%
  \let\@fs@iftopcapt\iftrue}
\renewcommand\fst@algorithm{\fs@ruled@notop}
\makeatother

\title{\large\bf Generating Grasp Poses for a High-DOF Gripper Using Neural Networks \vspace{-10px}}
\author{Min Liu$^{1,4}$, Zherong Pan$^{2}$, Kai Xu$^{1*}$, Kanishka Ganguly$^{3}$, and Dinesh Manocha$^{4}$  \\
\href{https://gamma.umd.edu/researchdirections/grasping/high_dof_grasping}{https://gamma.umd.edu/researchdirections/grasping/high\_dof\_grasping} \\
\vspace{-60px}
\thanks{$^{1}$Min Liu  and Kai Xu are with School of Computer, National University of Defense Technology. \{gfsliumin@gmail.com, kevin.kai.xu@gmail.com\}}
\thanks{$^{2}$Zherong is with Department of Computer Science, University of North Carolina at Chapel Hill. \{zherong@cs.unc.edu\}}
\thanks{$^{3}$Kanishka Ganguly is with UMIACS (Institute for Advanced Computer Studies), University of Maryland at College Park. \{kganguly@umiacs.umd.edu\}}
\thanks{$^{4}$Min Liu and Dinesh Manocha are with Department of Computer Science and Electrical \& Computer Engineering, University of Maryland at College Park. \{dm@cs.umd.edu\}}
\thanks{$^{*}$Kai Xu is the corresponding author.}}

\begin{document}
\maketitle
\thispagestyle{empty}
\pagestyle{empty}

\begin{abstract}
We present a learning-based method for representing grasp poses of a high-DOF hand using neural networks. Due to redundancy in such high-DOF grippers, there exists a large number of equally effective grasp poses for a given target object, making it difficult for the neural network to find consistent grasp poses. We resolve this ambiguity by generating an augmented dataset that covers many possible grasps for each target object and train our neural networks using a consistency loss function to identify a one-to-one mapping from objects to grasp poses. We further enhance the quality of neural-network-predicted grasp poses using a collision loss function to avoid penetrations. We use an object dataset that combines the BigBIRD Database, the KIT Database, the YCB Database, and the Grasp Dataset to show that our method can generate high-DOF grasp poses with higher accuracy than supervised learning baselines. The quality of the grasp poses is on par with the groundtruth poses in the dataset. In addition, our method is robust and can handle noisy object models such as those constructed from multi-view depth images, allowing our method to be implemented on a 25-DOF Shadow Hand hardware platform.
\end{abstract}
\begin{figure*}[h]
\vspace{0px}  
\begin{minipage}[h]{\linewidth}
\centering
\includegraphics[width=0.9\textwidth]{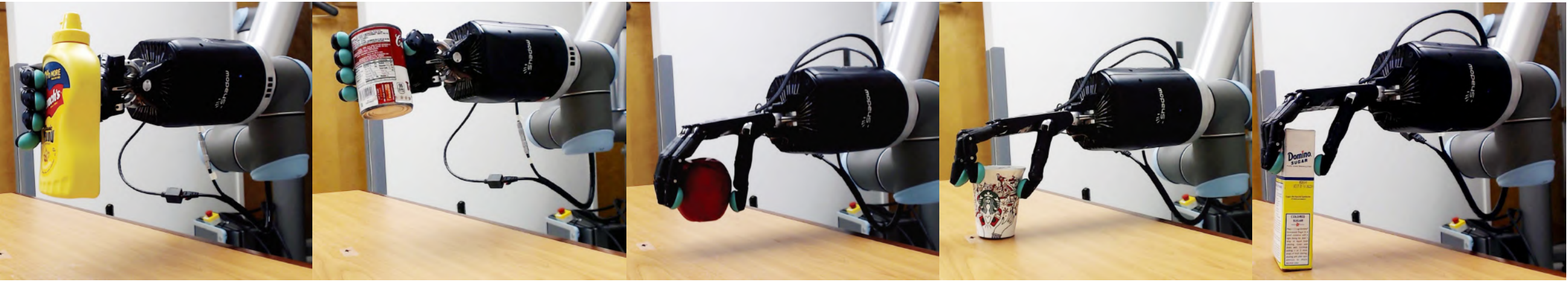}
\put(-230,-12){(a)}
\end{minipage}
\begin{minipage}[h]{\linewidth}
\vspace{5px}
\centering
\includegraphics[width=0.9\textwidth]{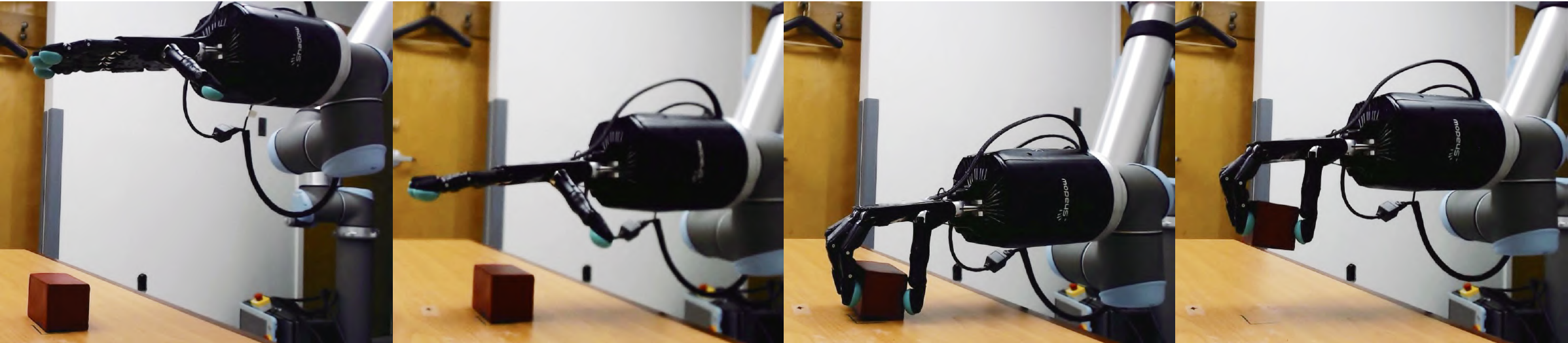}
\put(-230,-12){(b)}
\end{minipage}
\vspace{-3px}
\caption{(a): Our method implemented on the 25-DOF Shadow Hand to grasp different objects. (b): Several frames of a single grasp process.}
\vspace{-15px}
\label{fig:mulit_view_real_hardware}
\end{figure*}

\section{\label{intro}Introduction}
Grasp pose generation and prediction are important problems in robotics \cite{Xue2009AnAG,4755985,kopicki2016one}. Recently, learning-based methods \cite{mahler2019learning,doi:10.1177/0278364914549607,doi:10.1177/0278364917710318} have achieved high rates of success in terms of grasping unknown objects. However, these methods are mainly limited to low-DOF grippers with only 1-6 DOFs or they assume that a high-DOF gripper moves in a low-DOF subspace \cite{8304630}. This assumption limits the space of the grasp poses a robot hand can represent and the space of the target objects the hand can handle. In this work, we address the problem of developing learning algorithms for generating grasp poses for a high-DOF hand. Such high-DOF hands have been used to perform complex in-hand manipulations in prior works \cite{Andrychowicz2018LearningDI,Rajeswaran-RSS-18}. 

Generating grasp poses for a high-DOF gripper is more challenging than for low-DOF grippers due to the existence of pose ambiguity, i.e. there exists a large number of equally effective grasp poses for a given target object. However, we need to train a single neural network to predict one grasp pose for each object. As a result, we need to pick a set of grasp poses for a set of target objects, that can be represented by the neural network. In the case of a low-DOF gripper, if the neural network predicts the correct direction and orientation towards the object, one can simply close the gripper and the predicted grasp operation will very likely be successful. Therefore, most prior works \cite{mahler2019learning,DBLP:conf/icra/ZengSYDHBMTLRFA18,8304630} only learn the approaching direction and orientation of the gripper. For a high-DOF gripper, however, there are multiple remaining DOFs (beyond direction and orientation) to be determined after the wrist pose is known. Computing these remaining DOFs is still a major challenge in deep-learning-based grasp pose generation methods.

There are two kinds of learning-based methods for grasp pose generation. In the first \cite{doi:10.1177/0278364917710318,Fang2018MultiTaskDA}, a grasp pose is generated using two steps. First, a neural network is trained to predict the possibility of success given a grasp pose as an input. Second, the grasp pose is generated during runtime using a sampling-based optimizer such as a multi-armed bandit \cite{mahler2016dex} to maximize the possibility of success. This method does not suffer from pose ambiguity because it allows multiple grasps to be equally effective for a single object. However, the high-DOF nature of the gripper results in a large search space for the sampling-based optimizer, making the online phase very computationally costly. In the second kind of method \cite{8304630}, a neural network is trained to predict the grasp poses directly from single-view observations of the object. As a result, this direct method becomes very efficient because only a forward propagation through the neural network is needed to generate the grasp pose. However, since many high-DOF grasp poses can be equally effective for a single object, an additional constraint is required to guide neural networks to determine the poses from which it should learn. Due to the lack of such guidance, \cite{8304630} can not be used to generate high-DOF grasp poses directly.

\TE{Main Results:} We present a learning-based method for representing grasp poses for a high-DOF articulated robot hand. Our method enables fast grasp pose generation without the low-DOF assumption. Similar to \cite{8304630}, we train a neural network to predict grasp poses directly so that grasp poses can be generated efficiently during runtime. To resolve the ambiguity of grasp poses for each object, we introduce the notion of \TE{consistency loss}, which allows the neural network to choose from a large number of candidate grasp poses and select the one that can be consistently represented by a single neural network. However, the grasp pose predicted by the network can be in close proximity to the object, leading to many hand-object penetrations. To resolve this issue, we introduce collision loss, which penalizes any robot-object penetrations, to push the gripper outside the object. We train the neural network for $40$ hours on a dataset of $324$ objects by combining the BigBIRD Database \cite{singh2014bigbird}, the KIT Object Models Database \cite{kasper2012kit}, the YCB Benchmarks \cite{calli2015benchmarking}, and the Grasp Database \cite{2015_ICRA_kbs}. We show that our method can achieve $4\times$ higher accuracy (in terms of distances to the groundtruth grasp poses) than supervised learning baselines in grasp pose representation. In addition, we show that our method can be used in several application scenarios by taking inaccurate 3D object models as inputs; these object models are reconstructed from multi-view depth images. As a result, our method can be implemented on 25-DOF Shadow Hand hardware, as shown in \prettyref{fig:mulit_view_real_hardware}, where each grasp can be computed within 3-5 seconds at runtime.

The rest of the paper is organized as follows. We review related work in \prettyref{sec:related} and then formulate our problem in \prettyref{sec:problem}. The main neural network architecture and training algorithm are presented in \prettyref{sec:method}. Finally, we highlight the performance on different objects in \prettyref{sec:experi}.
\section{\label{sec:related}Related Work}
Methods for robot grasp pose generation can be classified based on the assumptions they make about the inputs. Early works \cite{6335488,772531,770384,Miller2004} are designed for complete 3D shapes, such as 3D triangulated meshes of objects, as inputs. To estimate the quality of a grasp pose \cite{6335488,772531} or compute a feasible motion plan \cite{770384,Miller2004} deterministically, a 3D mesh representation is used. However, these methods are difficult to deploy on current real-world grasping systems or robot hands due to discrepancies and sensing uncertainties. Most practical grasp planning methods that can take incomplete shapes are based on machine learning. Early learning methods predict good grasp poses \cite{Jiang2011EfficientGF} or points \cite{doi:10.1177/0278364907087172} from several RGB images using manually engineered features and supervised/active learning \cite{Montesano:2012:ALV:2109688.2109870}. More recently, learning-based grasp pose prediction algorithms \cite{mahler2016dex,mahler2019learning,doi:10.1177/0278364917710318,8304630,8460875,7759657,7487517} replace manually engineered features with features learned from deep convolutional networks for better generality and robustness. All these methods are designed for low-DOF grippers. Some learning-based methods \cite{1277,Yan2018Learning6G} take an incomplete or a partial shape as input and internally reconstruct a voxelized shape. Our method uses \cite{Miller2004} to generate groundtruth grasping data and we assume that the input to the neural-network is a complete object model represented using an occupancy grid. However, our trained network is robust to data inaccuracies and can be applied to object models reconstructed from multi-view depth images.

Most existing learning-based methods \cite{Jiang2011EfficientGF,mahler2016dex,mahler2019learning,doi:10.1177/0278364917710318,8304630,lu2017grasp} use the learned model in a two-stage algorithm. During the first step, the learned neural network takes both the observation of the object and a proposed grasp pose as inputs and predicts the possibility of a successful grasp. During the second step, the final grasp pose is optimized to maximize the rate of success using exhaustive search \cite{Jiang2011EfficientGF}, gradient-based optimization \cite{lu2017grasp,lu2019grasp}, sampling-based optimization \cite{doi:10.1177/0278364917710318}, or multi-armed bandits \cite{mahler2016dex}. Instead, our method uses a learning model to predict the grasp poses for a high-DOF gripper directly. Our method is similar to \cite{8304630}, which learns a neural network to predict the grasp poses directly, but \cite{8304630} is designed for low-DOF grippers and pose ambiguity is not handled.
\section{\label{sec:problem}Problem Formulation}
In this section, we formulate the problem of high-DOF grasp pose generation. Each grasp pose is identified with a high-DOF configuration of the robot hand $\E{x}=\TWOR{\E{x}_b}{\E{x}_j}$, where $\E{x}_b$ is the 7-DOF rigid transformation of the hand wrist and $\E{x}_j$ is the remaining DOFs, i.e., joint angles. Our goal is to find a mapping function $f(\E{o})=\E{x}$, where $\E{o}$ is an observation of the object $\mathcal{O}$. This observation can take several forms. In this paper, we assume that $\E{o}$ is the 3D occupancy grid \cite{choy20163d,Girdhar16b} derived by discretizing the object. We denote $\E{o}_s$ as the signed distance field \cite{osher2006level} derived by solving the Eikonal equation from the original mesh.

We use deep neural networks to represent $f$ with optimizable parameters denoted by $\boldsymbol{\theta}$. The main difference between our method and prior deep-learning-based methods \cite{Yan2018Learning6G,Fang2018MultiTaskDA,mahler2016dex} is that our network directly outputs the grasp pose $\E{x}$. Prior methods only predict the possibility of successful grasps, given a possible grasp pose, which can be summarized as a function $g(\E{x},\E{o})=p$, where $p$ is the possibility of success. Function $g$ has advantages over our function $f$ because $g$ allows multiple versions of $\E{x}$ to be generated for a single $\mathcal{O}$. However, to use $g$, we need to solve the following problem:
\begin{align}
\label{eq:search}
\argmax{\E{x}}\quad g(\E{x},\E{o}),
\end{align}
which can be computed efficiently for low-DOF grippers using either sampling-based optimization \cite{doi:10.1177/0278364917710318} or multi-armed bandits \cite{mahler2016dex}. However, this optimization can be computationally costly for a high-DOF gripper due to the high-dimensional search space. This optimization can also be ill-posed and under-determined because many unnatural grasp poses might also lead to effective grasp poses, as shown in \cite{4399227}. This is our main motivation for choosing $f$ over $g$. However, training a neural network that represents function $f$ is more challenging than training $g$ for two reasons.
\begin{itemize}
\item If we have a dataset of $N$ objects and groundtruth grasp poses $\{<\mathcal{O}_i,\E{x}_i>\}$, a simple training method is to use the data loss $\mathcal{L}_{data}=\sumno_i\|f(\E{o}_i,\boldsymbol{\theta})-\E{x}_i\|^2$. However, since multiple grasp poses $\E{x}_i$ are valid for each object $\mathcal{O}_i$, we can build many datasets for the same set of objects $\{\mathcal{O}_i\}$ by choosing different grasp poses for each object. The resulting data loss $\mathcal{L}_{data}$ generated by using different datasets can be considerably different according to our experiments. Therefore, the first challenge in training function $f$ is that we need to build a dataset leading to a small $\mathcal{L}_{data}$ after training.
\item A second problem in training $f$ is that we have to ensure the quality of grasp poses generated by the neural networks. The quality of a grasp pose in learning-based methods can be measured by comparing it with the groundtruth pose. However, there are other important metrics. For example, a grasp pose should not have penetration with $\mathcal{O}$. In prior methods \cite{Yan2018Learning6G,Fang2018MultiTaskDA,mahler2016dex}, the neural network is not responsible for ensuring the quality of the grasp poses, but we can guarantee high-quality grasp poses when solving \prettyref{eq:search} after training. However, in our case, the neural network is used to generate $\E{x}$ directly, so our final results are very sensitive to the outputs of the neural network.
\end{itemize}
\section{\label{sec:method}Learning High-DOF grasp poses}
In this section, we present the architecture of our neural network used for high-DOF grasping.
\begin{figure*}[t]
\centering
\vspace{0px}
\includegraphics[width=0.9\textwidth]{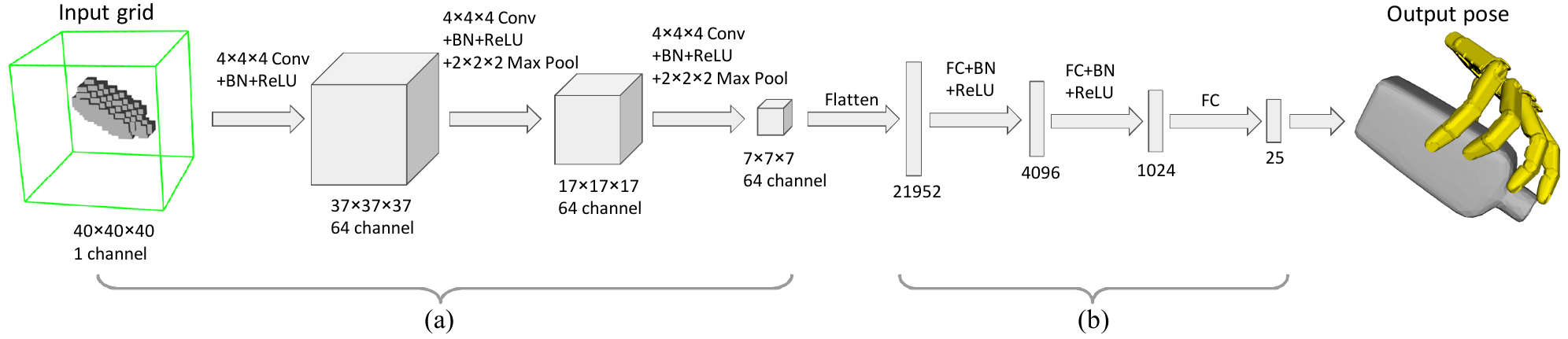}
\vspace{-10px}
\caption{An illustration of our two sub-networks. (a): $\E{NN}_\E{o}$ maps the observations of the object to the feature vector $\boldsymbol{\omega}$. (b): $\E{NN}_\E{x}$ maps the feature vector $\boldsymbol{\omega}$ to the grasp pose $\E{x}$. We use the same network architecture for different robot hand hardwares with different DOFs by modifying only the output layer.}
\vspace{-10px}
\label{fig:NN}
\end{figure*}

\subsection{Neural Networks}
We represent $f$ using a deep neural network, as illustrated in \prettyref{fig:NN}. We assume that a high-DOF grasp pose can be generated from a low-dimensional feature vector of the object denoted by $\boldsymbol{\omega}$; a similar approach is used by \cite{8304630}. We use a fully connected sub-network $\E{NN}_\E{x}$ to parameterize this mapping function:
\begin{align*}
\E{x}=\E{NN}_\E{x}(\boldsymbol{\omega},\theta_1),
\end{align*}
where $\theta_1$ is the optimizable weights. To parameterize $\E{NN}_\E{x}$, we use a network with 3 hidden layers with $\left(64\times7\times7\times7=\right)21952, 4096, 1024$ neurons, respectively. We use ReLU activation functions for each hidden layer and we add batch normalization to the first two hidden layers. When different sensors leading to different observations of $\mathcal{O}$ (e.g., an occupancy grid or a depth image,) are used in our application, we use another sub-network to transform the observation to $\boldsymbol{\omega}$. Therefore, we have:
\begin{align*}
\E{NN}_\E{o}(\E{o},\theta_2)=\boldsymbol{\omega},
\end{align*}
and $\theta\triangleq\TWOR{\theta_1}{\theta_2}$. This neural network is fully convolutional. $\E{NN}_\E{o}$ has 3 3D-convolutional layers with 64 kernels of size 4. We add batch normalization, ReLU activation, and max-pooling layers after each convolutional layer. Finally, we have $f=\E{NN}_\E{x}\circ\E{NN}_\E{o}$.

\subsection{Consistency Loss}
In practice, optimizing $\theta$ is difficult due to the two challenges discussed in \prettyref{sec:problem}. We resolve these issues using two loss functions. Our first loss function is called a consistency loss function and we use this function used to resolve the grasp pose ambiguity for each $\mathcal{O}$. Instead of picking one grasp pose $\E{x}_i$ for each $\mathcal{O}_i$ during dataset construction, we compute a set of $K$ grasp poses denoted by $\E{x}_{i,j}$ for each $\mathcal{O}_i$, where $j=1,\cdots,K$, resulting in a large dataset with $NK$ grasp poses for $N$ objects. As a result, our consistency loss function takes the following form: 
\begin{align*}
\mathcal{L}_{consistency}=\sumno_i\fmin{j}\|f(\E{o}_i,\boldsymbol{\theta})-\E{x}_{i,j}\|^2/N.
\end{align*}
This novel formulation allows the neural network to pick the $N$ grasp poses leading to the smallest residual. Note that, although $\mathcal{L}_{consistency}$ is not uniformly differentiable, its sub-gradient exists and optimizing $\mathcal{L}_{consistency}$ with respect to both $\theta$ and $j$ can be performed with the conventional back-propagation gradient computation framework \cite{ketkar2017introduction}. Specifically, after forward propagation computes $f(\E{o}_i,\boldsymbol{\theta})$ for every $i$, we pick $j$ leading to the smallest residual, and finally perform backward propagation with:
\begin{align*}
&\FPP{\mathcal{L}_{consistency}}{f(\E{o}_i,\boldsymbol{\theta})}=
(f(\E{o}_i,\boldsymbol{\theta})-\E{x}_{i,j^*})/N \\
&j^*=\argmin{j}\|f(\E{o}_i,\boldsymbol{\theta})-\E{x}_{i,j}\|^2.
\end{align*}

\subsection{Collision Loss}
To resolve the second challenge and ensure the quality of the learned grasp poses, we note that most incorrect or inaccurate $\E{x}$ predicted by the neural network have the gripper intersecting with $\mathcal{O}$. To resolve this problem, we add a second loss function that penalizes any penetrations between the gripper and $\mathcal{O}$. Specifically, we first construct a signed distance function $\E{o}_s$ from the original mesh and then sample a set of points $\E{p}_{i=1,\cdots,P}$ on the gripper. Next, we formulate the collision loss function as:
\begin{align*}
\mathcal{L}_{collision}=\sumno_{i=1}^P\E{min}^2(\E{o}_s(\E{T}(\E{p}_i,f(\E{o}_i,\boldsymbol{\theta}))),0),
\end{align*}
where $\E{T}$ is the forward kinematics function of the gripper transforming $\E{p}_i$ to its global coordinates. We also assume $\E{o}_s$ has positive values outside $\mathcal{O}$ and negative values inside $\mathcal{O}$. Again, $\mathcal{L}_{collision}$ is not uniformly differentiable but has a well-defined sub-gradient, so it can be used to optimize the neural network. In our experiments, we find that the quality of the learned grasp poses is sensitive to the selection of sample points $\E{p}_i$. We choose to use the same set of sample points for dataset generation and the collision loss function. Specifically, we use simulated annealing \cite{4399227} to generate groundtruth grasp poses. \cite{4399227} optimizes an approximate grasp quality function that measures the distance between a set of desired contact points to the object surfaces. These contact points are also used as sample points in $\mathcal{L}_{collision}$.

\subsection{Combined Loss}
The consistency loss and the collision loss are combined using parameters $\beta$ as shown in the following equation:
\begin{align}
\label{eq:combined}
\mathcal{L}_{combined}=\beta*\mathcal{L}_{consistency} + (1-\beta)*\mathcal{L}_{collision},
\end{align}
where the relative weight $\beta$ is between 0 and 1. Empirically, we find that grasp results of higher values $\beta$ are more like \textit{GraspIt!} groundtruth, while results of lower values $\beta$ bring the fingers closer to the surfaces of objects, which in turn results in higher success rates.

\subsection{Pose Refinement}
After a neural network predicts a nominal grasp pose for an unknown object, we can further refine it at runtime by looking for another grasp pose that is close to the nominal pose but does not have any intersections with the object. To do this, we solve a simple optimization. Specifically, after the neural network predicts $f(\E{o},\theta)$, we first search for another pose $\E{x}^*$ closest to it by minimizing the following objective function:
\begin{align*}
\argmin{\E{x}^*}\quad\beta*\mathcal\|\E{x}^*-f(\E{o},\theta)\|^2 + (1-\beta)*\mathcal{L}_{collision}.
\end{align*} 
We call this procedure pose refinement.

\section{\label{sec:experi}Implementation and Performance}
In this section, we provide more details about our experiment platform setup, results, and evaluations.

\subsection{Grasp Training Dataset Generation}
Given a set of target objects, we take three steps to generate our grasp pose training dataset. First, we use an existing sampling-based motion planner, \GRASPIT{ }\cite{Miller2004}, to generate many high-quality grasp poses for each object. We then perform data augmentation via global rigid transformation. Finally, we compute a signed distance field for each of the target objects. 

\setlength{\columnsep}{6pt}
\begin{wrapfigure}{r}{5cm}
\includegraphics[width=5cm]{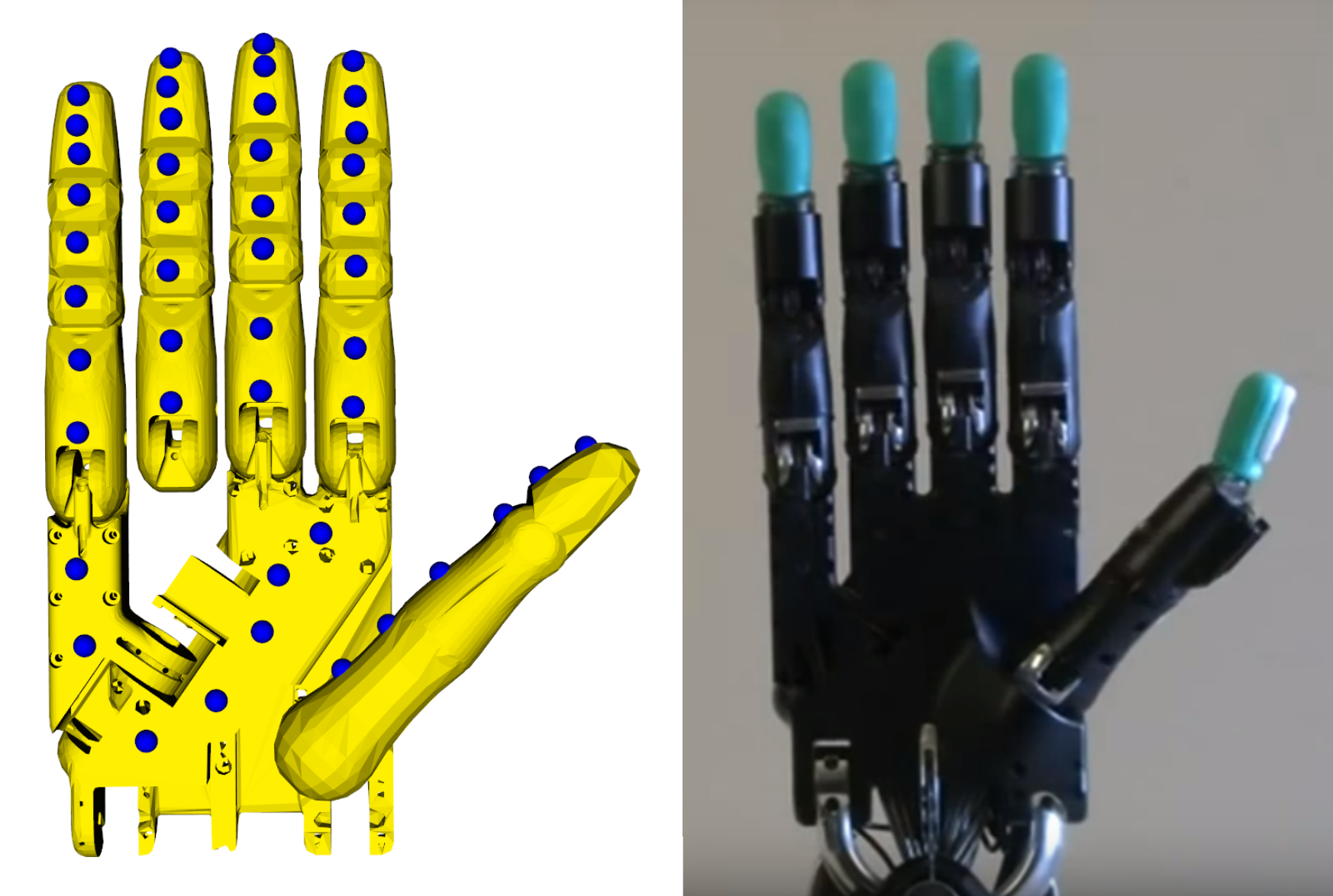}
\caption{Left: The Shadow Hand model we use and the sampled potential contact points in blue. Right: The real Shadow Hand.}
\vspace{-10px}
\label{fig:shadowhand}
\end{wrapfigure}
\subsubsection{Grasp Pose Generation} 
\begin{figure}[h]
\centering
\vspace{-5px}
\includegraphics[width=0.45\textwidth]{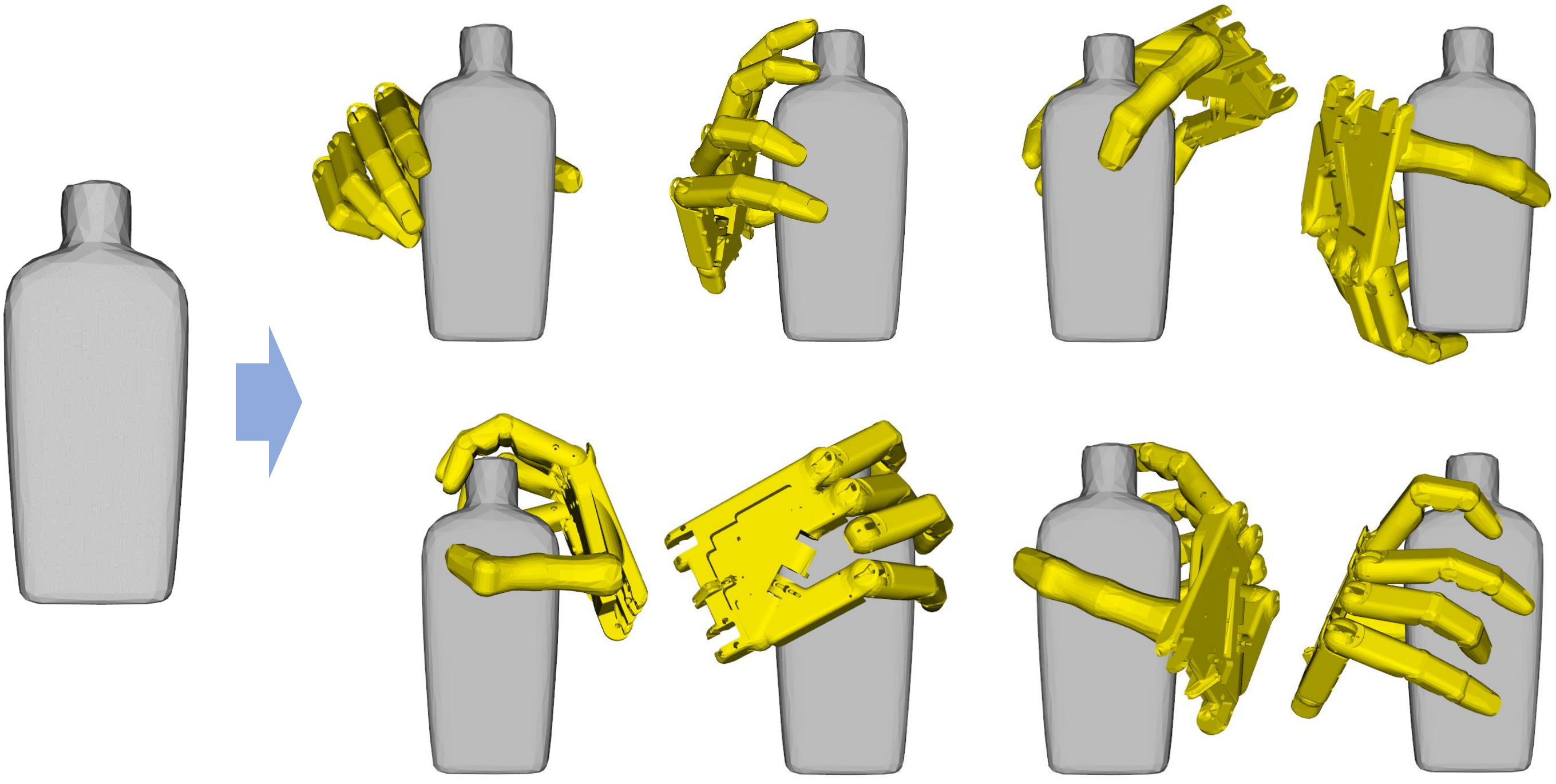}
\vspace{0px}
\caption{An illustration of some sample grasp poses (yellow) for a single object (gray).}
\vspace{-8px}
\label{fig:grasp_gen}
\end{figure}

We collect object models from several datasets, including BigBIRD \cite{singh2014bigbird}, KIT Object Models Database \cite{kasper2012kit}, YCB Benchmarks \cite{calli2015benchmarking}, and Grasp Database \cite{2015_ICRA_kbs}. Our dataset contains $N=324$ mesh models, of which most are everyday objects. Our high-DOF gripper is the Shadow Hand with 25 DOFs, as shown in \prettyref{fig:shadowhand}. Given an initial pose of the Shadow Hand, \GRASPIT { }uses an optimization-based planner to find an optimal grasp pose that minimizes a cost function, which is found via simulated annealing. The cost function can take various forms and we use the sum of distances between sample points and object surfaces as the cost function. We run simulated annealing for 10000 iterations, where the planner generates and evaluates 10 candidate grasp poses during each iteration. To generate many redundant grasp poses for each object, we run the simulated annealing algorithm for $K=100$ times from random initial poses. Altogether, the groundtruth grasp pose dataset is generated by calling the simulated annealing planner $324\times100$ times. Some grasp poses for an object are illustrated in \prettyref{fig:grasp_gen}.

\subsubsection{Data Augmentation}
\begin{figure}[h]
\centering
\vspace{-8px}
\includegraphics[width=0.45\textwidth]{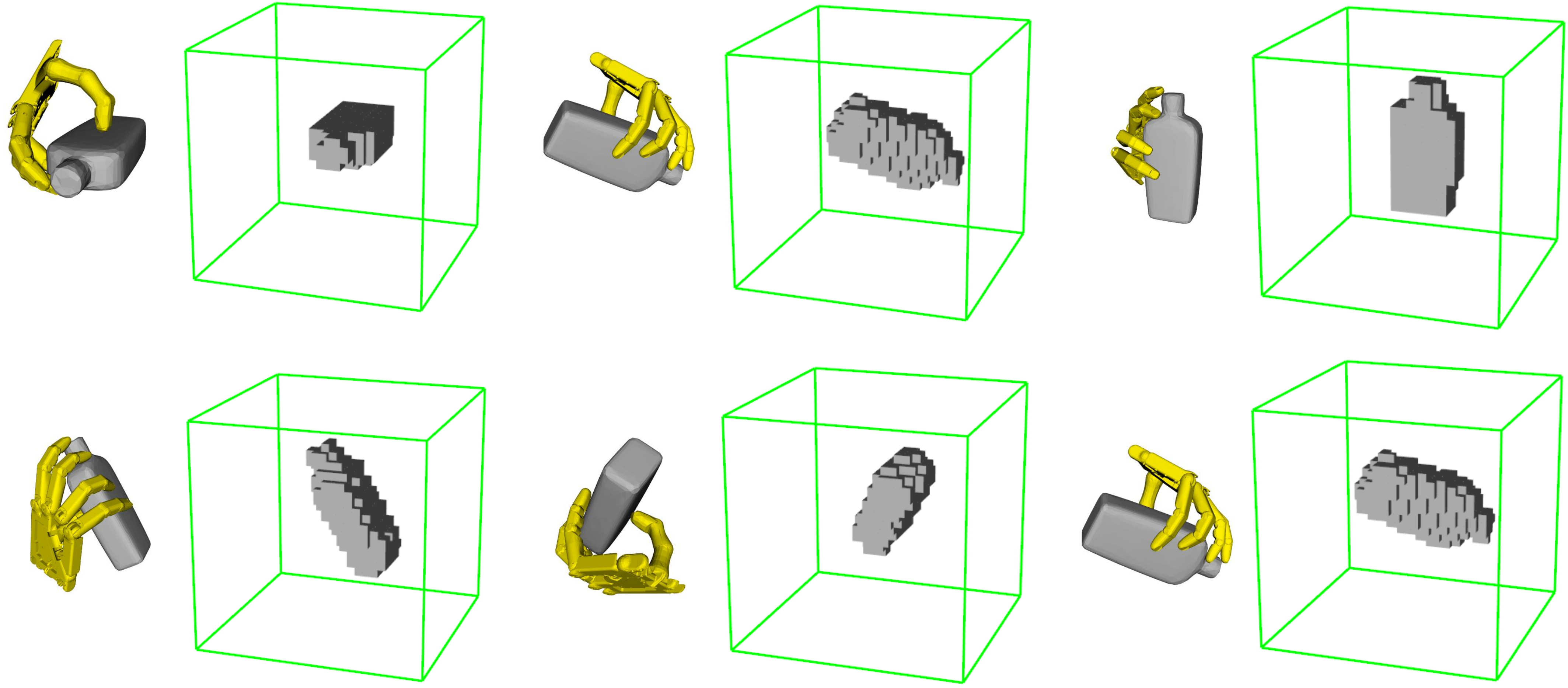}
\vspace{0px}
\caption{An illustration of rotated object poses and grasp poses generated by data augmentation.}
\vspace{-10px}
\label{fig:data_aug}
\end{figure}

Generating groundtruth grasp poses using a motion planner is very computationally costly, so we use a simple method to synthesize more data. The input to $\E{NN}_\E{v}$ is a voxelized occupancy grid. We move the objects, put their centers of mass at the origin of the Cartesian coordinate system, and then rotate each object along with its 100 best grasp poses along 27 different rotation angles and axes. These 27 rotations are derived by concatenating rotations along X, Y and Z-axes for  $60^\circ, 120^\circ, 180^\circ$, as illustrated in \prettyref{fig:data_aug}. For each rotation, we record the affine transformation matrix $\E{T}_{r}$. In this way, we generate a dataset that is 27 times larger than the original dataset. This data augmentation not only helps resist over-fitting when training neural networks but also helps make the neural network invariant to target object poses.

\subsubsection{Signed Distance Fields Construction} 
\begin{figure}[h]
\centering
\vspace{-5px}
\includegraphics[width=0.45\textwidth]{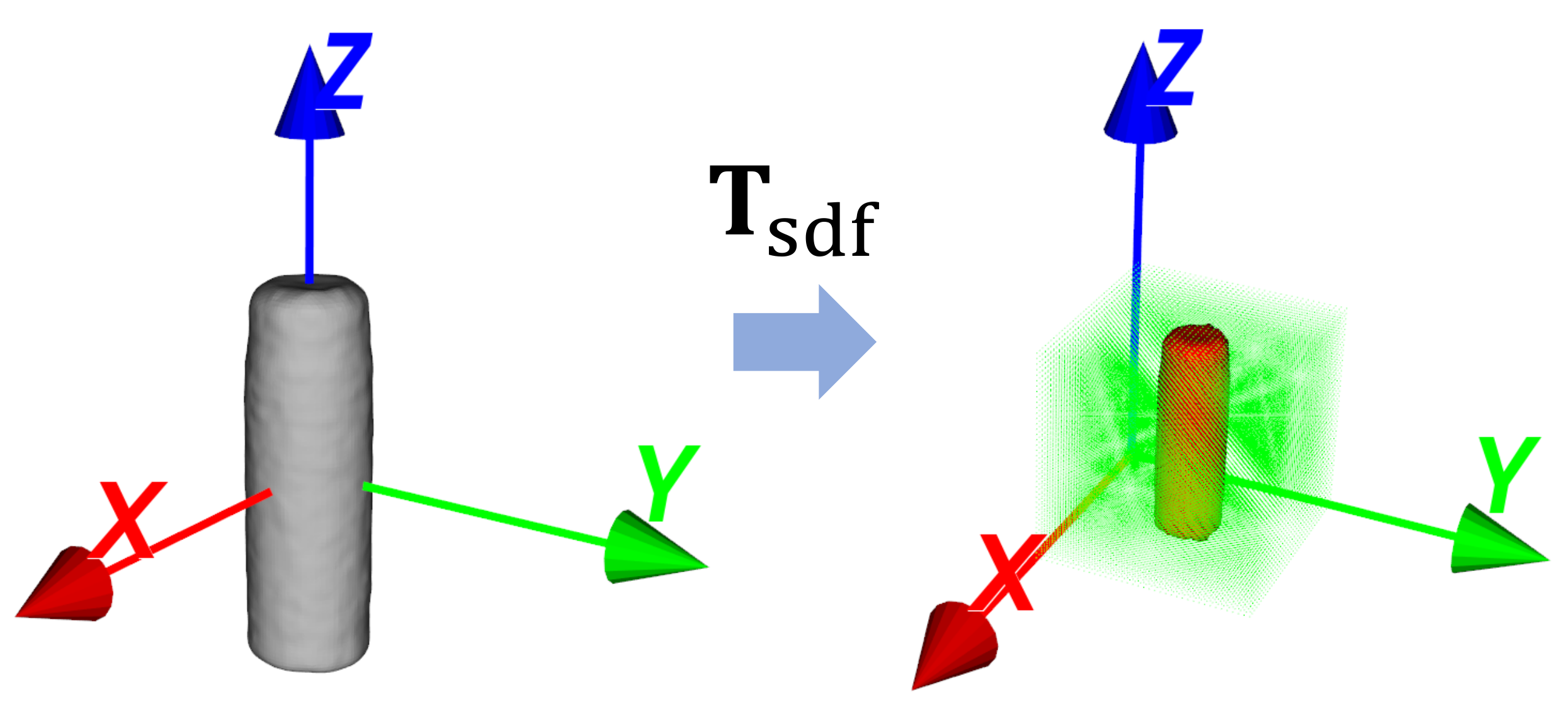}
\vspace{0px}
\caption{Signed distance fields construction, where the green area is the domain of $\E{o}_s$.}
\vspace{-10px}
\label{fig:sdf_transform}
\end{figure}

To calculate the collision loss when training our neural networks, we compute a signed distance field $\E{G}_{sdf}$ for each target object by solving the Eikonal equation. We set the resolution of $\E{G}_{sdf}$ to $128^3$ and $\E{G}_{sdf}$ has a local coordinate system where $\E{G}_{sdf}$ occupies the unit cube between $[0,0,0]$ and $[1,1,1]$. If the maximal length of the object's bounding box is $L$, the transformation matrix from an object's local coordinate system to $\E{G}_{sdf}$'s local coordinate system is:
\begin{align*}
\E{T}_{sdf}=\begin{bmatrix}
 s&  0&  0& 0.5 \\ 
 0&  s&  0& 0.5\\ 
 0&  0&  s& 0.5\\ 
 0&  0&  0& 1
\end{bmatrix}\quad s\triangleq0.95/L,
\end{align*}
which is illustrated in \prettyref{fig:sdf_transform}.

\subsection{Experimental setup}
\begin{figure*}[ht]
\centering
\vspace{0px}
\includegraphics[width=0.9\textwidth]{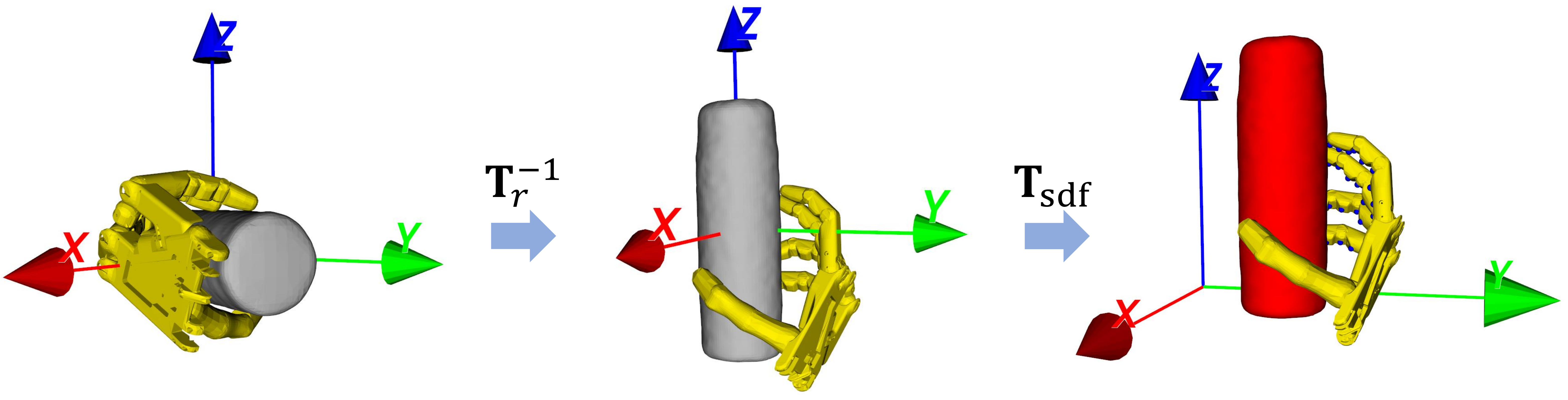}
\vspace{-10px}
\caption{To fit the size and place of the object to the size of the Shadow Hand, we use a combined transformation from the global coordinate system to $\E{G}_{sdf}$'s local coordinate system.}
\vspace{-15px}
\label{fig:fkw}
\end{figure*}

We split the set of 324 target objects into an $80\%$ (259) training set and a $20\%$ (65) testing set. Note that each object mesh is augmented to 27 meshes with different $\E{T}_{r}$. After we voxelize the meshes into 3D occupied grids, we get a total of $324\times27=8748$ grids, each with a related $\E{T}_{r}$. All augmented meshes related to the same object share the same $\E{G}_{sdf}$. Transformation from augmented meshes to $\E{G}_{sdf}$ is given as $\mathbf{T}_{sdf}\cdot \mathbf{T}_{r}^{-1}$. On the Shadow Hand, we sample $P=45$ potential contact points, as shown in \prettyref{fig:shadowhand}. To sample the signed distance field using $\E{T}(\E{p}_i,f(\E{o}_i,\boldsymbol{\theta}))$, we need to transform the point from the global coordinate system to the coordinate system of the signed distance field, which is:
\begin{align}
\label{eq:fwk}
\mathbf{T}_{sdf}\cdot \mathbf{T}_{r}^{-1}\E{T}(\E{p}_i,f(\E{o}_i,\boldsymbol{\theta})),
\end{align}
as shown in \prettyref{fig:fkw}. All experiments are carried out on a desktop with an Intel$^{\circledR}$ Xeon W-2123 @ 3.60GHz $\times$ 4, 32GB RAM, and an NVIDIA$^{\circledR}$ Titan Xp graphics card with 12GB memory, on which training the neural networks takes $40$ hours.

\subsection{Results and evaluation}
In this section, we evaluate the performance of our novel training method and demonstrate its benefits method for solving high-DOF grasp problems.

\subsubsection{Challenge of High-DOF Grasp Problems}
In our first benchmark, we highlight the challenges of dealing with high-DOF grippers and the necessity of our novel loss function for solving the problem. We first train our neural network using conventional supervised learning. In other words, we create a small dataset with each object corresponding to only one grasp pose ($K=1$), and we use the simple $L_2$ loss function:
\begin{align*}
\mathcal{L}_2=\sumno_i\|f(\E{o}_i,\boldsymbol{\theta})-\E{x}_{i}\|^2/N.
\end{align*}
With this loss function, we train two neural networks to represent grasp poses for both a high-DOF gripper (25-DOF Shadow Hand) and a low-DOF gripper (11-DOF Barrett Hand) and compare the residual of $\mathcal{L}_2$ after training. Due to pose ambiguity, supervised learning using the $L_2$ loss function can lead to inconsistency problems. Our experimental results in \prettyref{table:low_dof} also show that this inconsistency problem is more serious in high-DOF grippers. These two neural networks are trained using the ADAM algorithm \cite{kingma2014adam} with a fixed learning rate of 0.001, a momentum of 0.9, and a batch size of 16.
\begin{table}[h]
\centering
\begin{tabular}{|c|c|c|c|}
\hline
Hand    & DOFs of Grippers & 
\TWORCell{Residual of $\mathcal{L}_2$}{on Test Set} & 
\TWORCell{Residual of $\mathcal{L}_2$}{on Training Set} \\ \hline
Shadow  & 25 & 73.61 & 76.02    \\ \hline
Barrett & 11 & 5.84  & 4.76     \\ \hline
\end{tabular}
\caption{We train two neural networks using an $L_2$ loss function to represent grasp poses for the Shadow Hand and the Barrett Hand. The residual is much higher for the Shadow Hand on both the training set and the test set, meaning that high-DOF grippers suffer more from the inconsistency problem. This can be resolved using the consistency loss function.}
\label{table:low_dof}
\vspace{-10px}
\end{table}

\subsubsection{Consistency and Collision Loss}
As shown in \prettyref{table:shadow}, we train the neural network using our large dataset with $K=100$. In this experiment, we train three neural networks using two different loss functions, $\mathcal{L}_{consistency}$ and $\mathcal{L}_{combined}$, where we pick $\beta=0.75$. We have tried multiple choices of $\beta$ and found that 0.75 leads to the best results. After training each neural network, we evaluate it on the test set and summarize the residuals of different losses, leading to $6$ values in \prettyref{table:shadow}; we also copy the first row of \prettyref{table:low_dof} to \prettyref{table:shadow} as a reference for simple supervised learning method. Note that $\mathcal{L}_{consistency}$ and $\mathcal{L}_2$ both represent the distance from the neural-network-predicted grasp pose to a certain groundtruth pose, the only difference is that we have only one groundtruth pose in $\mathcal{L}_2$ and we have $K$ groundtruth poses in $\mathcal{L}_{consistency}$, so $\mathcal{L}_{consistency}$ and $\mathcal{L}_2$ are comparable.

From the first row of \prettyref{table:shadow}, we can see that, even when simple supervised learning is used at training time, the residual of $\mathcal{L}_{consistency}$ (2.630) is already much smaller than the residual of $\mathcal{L}_2$ (73.61). This means that the distance between the neural-network-predicted grasp pose and the closest groundtruth pose is much smaller than the average distance to all the 100 candidate grasp poses. If $\mathcal{L}_{consistency}$ is used as a loss function during training time, the residual of $\mathcal{L}_{consistency}$ is further reduced from $2.630$ to $0.043$. However, using combined loss does not further reduce residual metrics. In the next section, we will see that collision loss will result in grasp poses that are closer to object surfaces, which increases the success rate of grasping.
\begin{table}[h]
\centering
\begin{tabular}{|c|c|c|c|}
\hline
\backslashbox{Loss}{Residual} & 
$\mathcal{L}_2$ & 
$\mathcal{L}_{consistency}$                   & \TWORCell{$\mathcal{L}_{combined}$}
                                                {($\beta=0.75$)}                          \\ \hline
$\mathcal{L}_2$                               &   73.61   &   2.630    &  55.865          \\ \hline
$\mathcal{L}_{consistency}$                   &   0.914   &   0.043    &  0.261           \\ \hline
\TWORCell{$\mathcal{L}_{combined}$}
{($\beta=0.75$)}                              &   0.345   &  0.062     &  0.133           \\ \hline
\end{tabular}
\caption{We train neural networks using 3 different loss functions (different rows). After training, we summarize the residuals of different loss functions on the test set (different columns). Our consistency loss function drastically reduces the error of neural networks in representing a single grasp pose.}
\label{table:shadow}
\vspace{-10px}
\end{table}

\begin{figure*}[h]
\centering
\vspace{0px}
\includegraphics[width=0.9\textwidth]{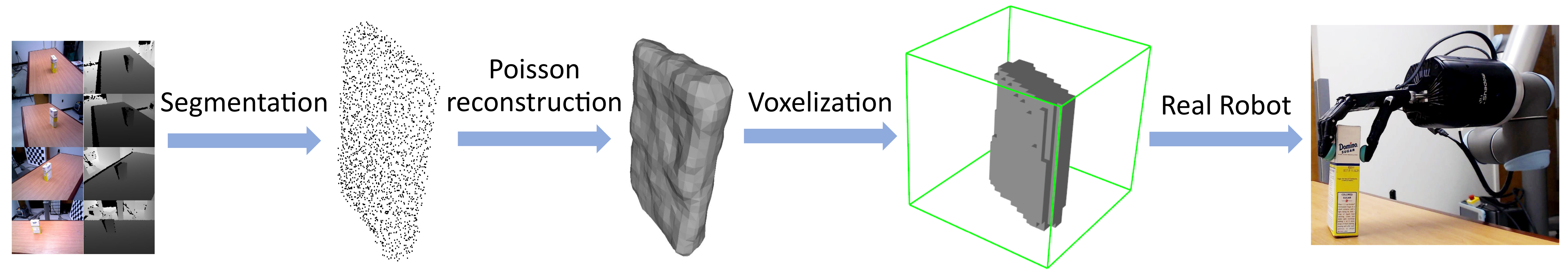}
\vspace{-3px}
\caption{Our neural network can use inaccurate object models reconstructed from multi-view depth images. The object meshes are reconstructed by first segmenting the point cloud and excluding the background, then applying Poisson surface reconstruction, and finally voxelizing the model.}
\vspace{-15px}
\label{fig:mulit_view}
\end{figure*}

\subsubsection{Penetration Handling}
\begin{figure}[ht]
\centering
\vspace{-5px}
\includegraphics[width=0.45\textwidth]{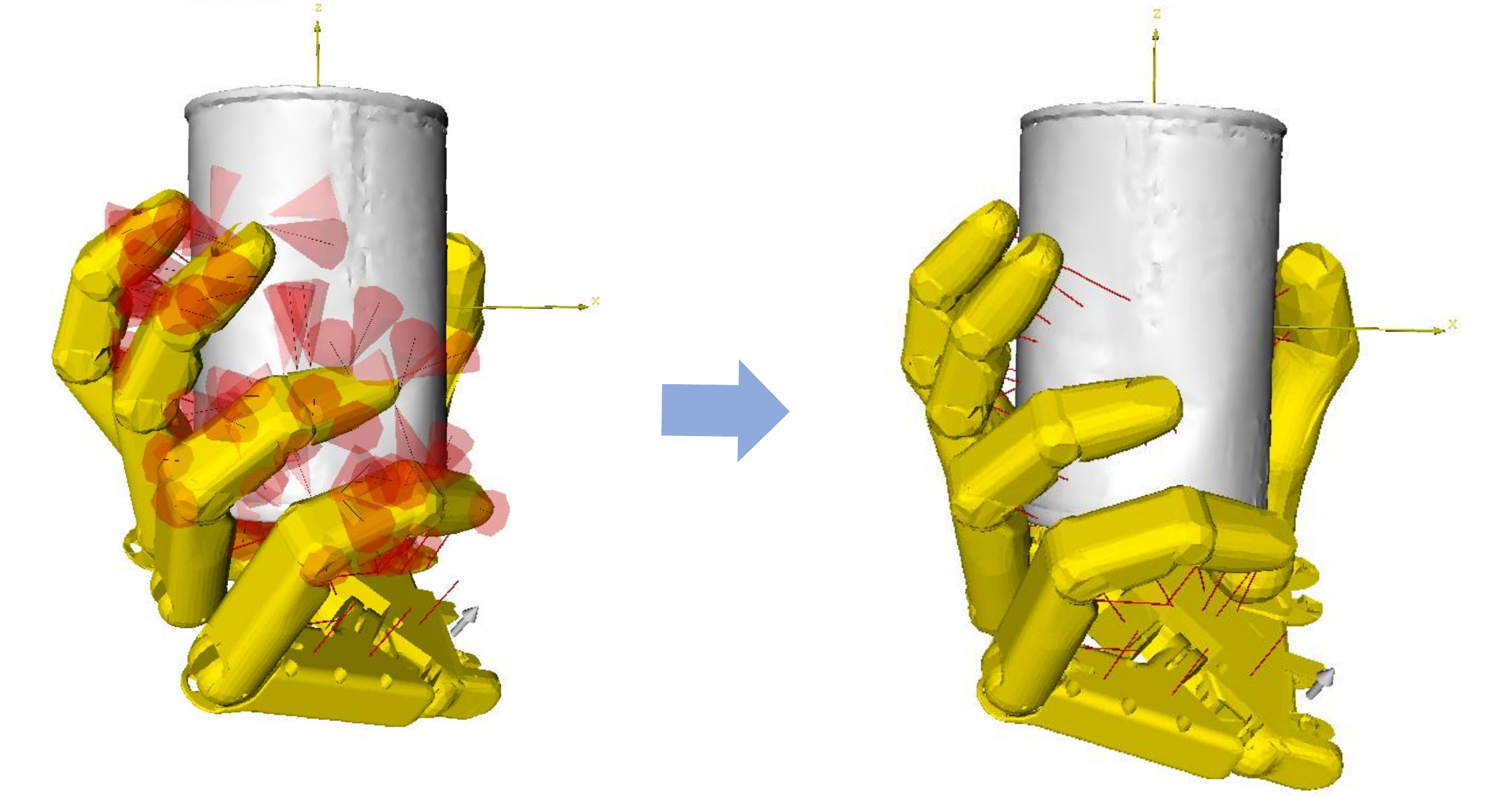}
\vspace{-5px}
\caption{There are still some penetrations after $\mathcal{L}_{collision}$ is  used (a), and we can resolve these penetrations (b) by using pose refinement during runtime.}
\vspace{-5px}
\label{fig:pr}
\end{figure}

Given an object, we first use the neural network to compute a proposed grasp pose. However, this grasp pose can be invalid and may have some penetrations into the target object. We can fix this problem by combining two methods. The first method is introducing collision loss at training time. From \prettyref{table:shadow}, we can see that introducing collision loss does not improve different residuals in general. However, it is very efficient in resolving most penetrations. In our experiment, introducing collision loss leads to an average relative change of the learned grasp pose by:
\begin{align*}
\frac{\|f_{+collision}(\E{o}_i,\boldsymbol{\theta})-f_{-collision}(\E{o}_i,\boldsymbol{\theta})\|}
{\|\E{x}_{i,j}\|}=12.7\%.
\end{align*}
When testing the neural network trained without collision loss on the test set, an average of $2.563$ of the $45$ sample points have penetrations with the target object on average and the penetration depth is $0.0553m$. With collision loss, the average number of sample points with penetration is reduced to $0.719$ and the average penetration depth is reduced to $0.0081m$. However, there are still some small penetrations, as shown in \prettyref{fig:pr} (a). During runtime, the actual grasping hardware cannot allow any penetrations between the object and the Shadow Hand. A second method is needed to compute a grasp pose without any penetrations; we use a simple interpolation method (runtime adjustment). Specifically, we compute the gradient of $\mathcal{L}_{collision}$ with respect to the joint angles:
\begin{align*}
\FPP{\left[\sumno_{i=1}^P\E{min}^2(\E{o}_s(\E{T}(\E{p}_i,\E{x})),0)\right]}{\E{x}}
\end{align*}
and we update our joint pose along the negative gradient direction until there are no penetrations. In practice, a single forward propagation through the neural network takes a computational time of $0.541s$ and the runtime adjustment takes $0.468s$.

\subsubsection{Multi-View Depth Image as Input}
\begin{figure}[h]
\centering
\vspace{-10px}
\includegraphics[width=0.45\textwidth]{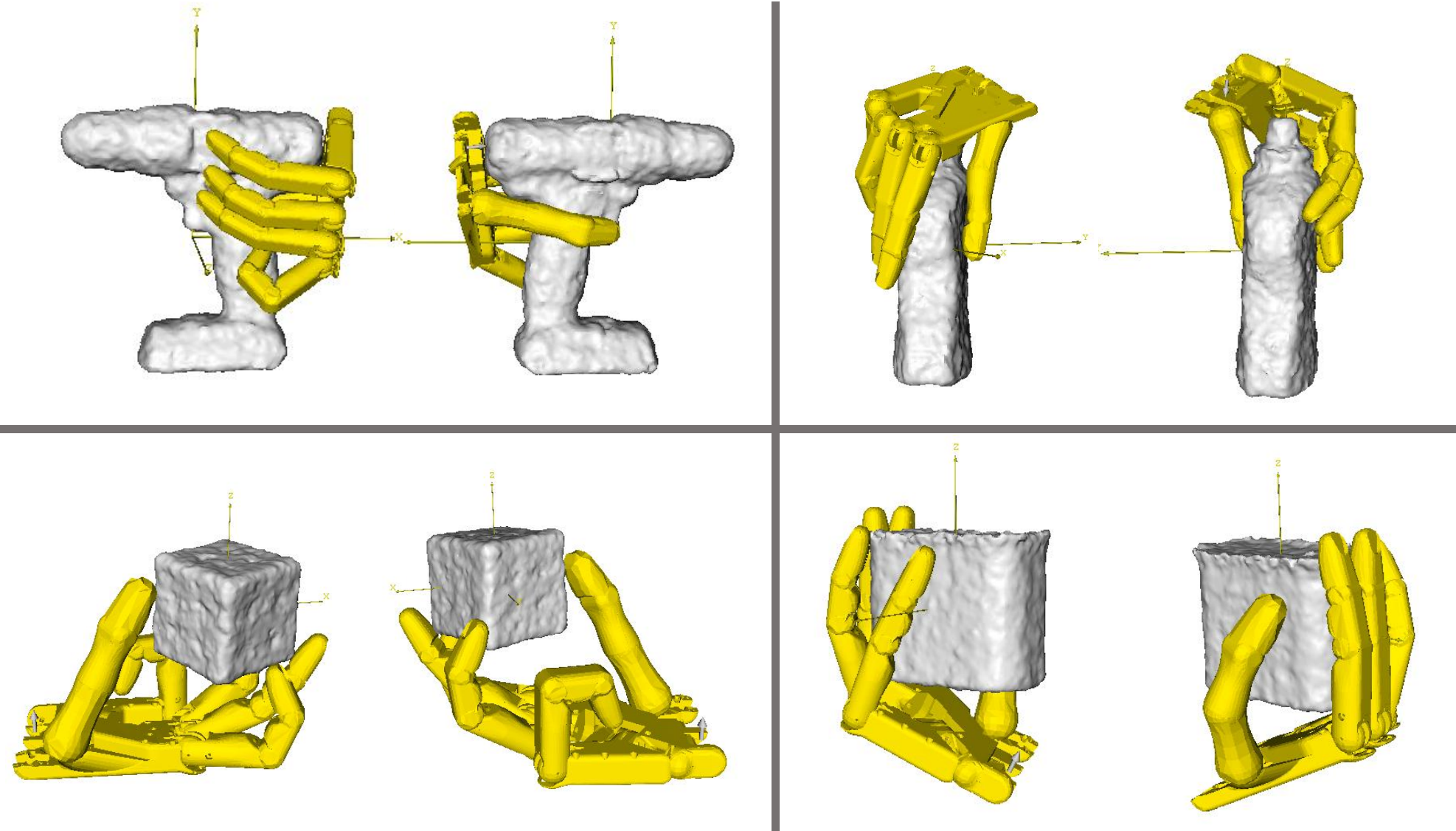}
\vspace{-5px}
\caption{High-quality grasp poses from 2 different views for 4 objects.}
\vspace{-5px}
\label{fig:multi_view_result}
\end{figure}

To extend our method to real-world scenarios, we evaluate our method on object models with uncertainties or inaccuracies. The real Shadow Hand is mounted on a UR10 arm and we use Shadow Robot Interface to move the arm and hand. In our experiment, we select 15 objects from the YCB Benchmarks, none of which have been included in our training dataset or test dataset. These 15 objects are captured using a multi-view depth camera and their geometric shapes are constructed using the standard pipeline implemented in \cite{5980567} and illustrated in \prettyref{fig:mulit_view}. Specifically, RANSAC is first used to remove the planar background of the obtained point cloud, Euclidean cluster extraction algorithm is used to find a set of segmented object point clouds, and segmented object meshes are then extracted using Poisson surface reconstruction. The reconstructed meshes are finally voxelized to a 3D occupancy grid. After pre-process, reconstructed mesh is fed to our neural networks to generate grasp poses. Sometimes the generated grasp poses are of low quality in terms of the $\epsilon$-metric, in which case we rotate the object mesh and run our neural networks again to generate a new grasp pose. On average, we rotate the object 3-5 times and report the best grasp pose quality in the wrench space. Although these reconstructed object meshes have noisy surfaces, we still get an average grasp quality of 0.102 over the 15 objects where we use the $\epsilon$-metric to measure grasp quality. Some grasp poses are shown in \prettyref{fig:multi_view_result}. We carry out our real robot grasping for 15 objects after finding a feasible grasp for each and 12 are successful.

\begin{figure}[h]
\centering
\vspace{0px}
\includegraphics[width=0.5\textwidth]{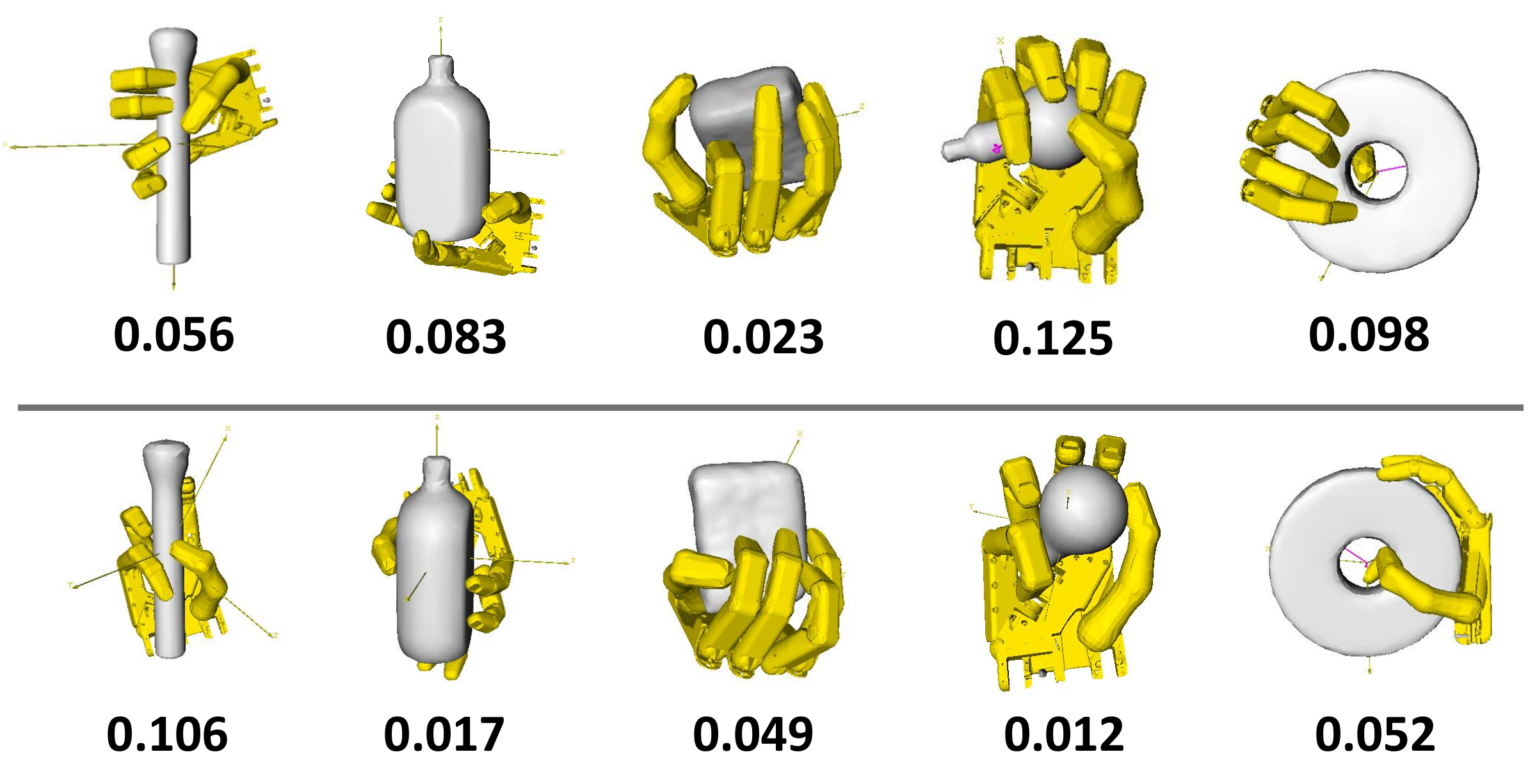}
\vspace{-10px}
\caption{A comparison of grasp pose quality generated using \GRASPIT{ }(top row) and our method (bottom row).}
\vspace{-10px}
\label{fig:comparison}
\end{figure}
\subsubsection{\label{subsubsec:grasp_quality}Comparison with Prior Methods}
\begin{figure}[h]
\centering
\vspace{0px}
\includegraphics[width=0.45\textwidth]{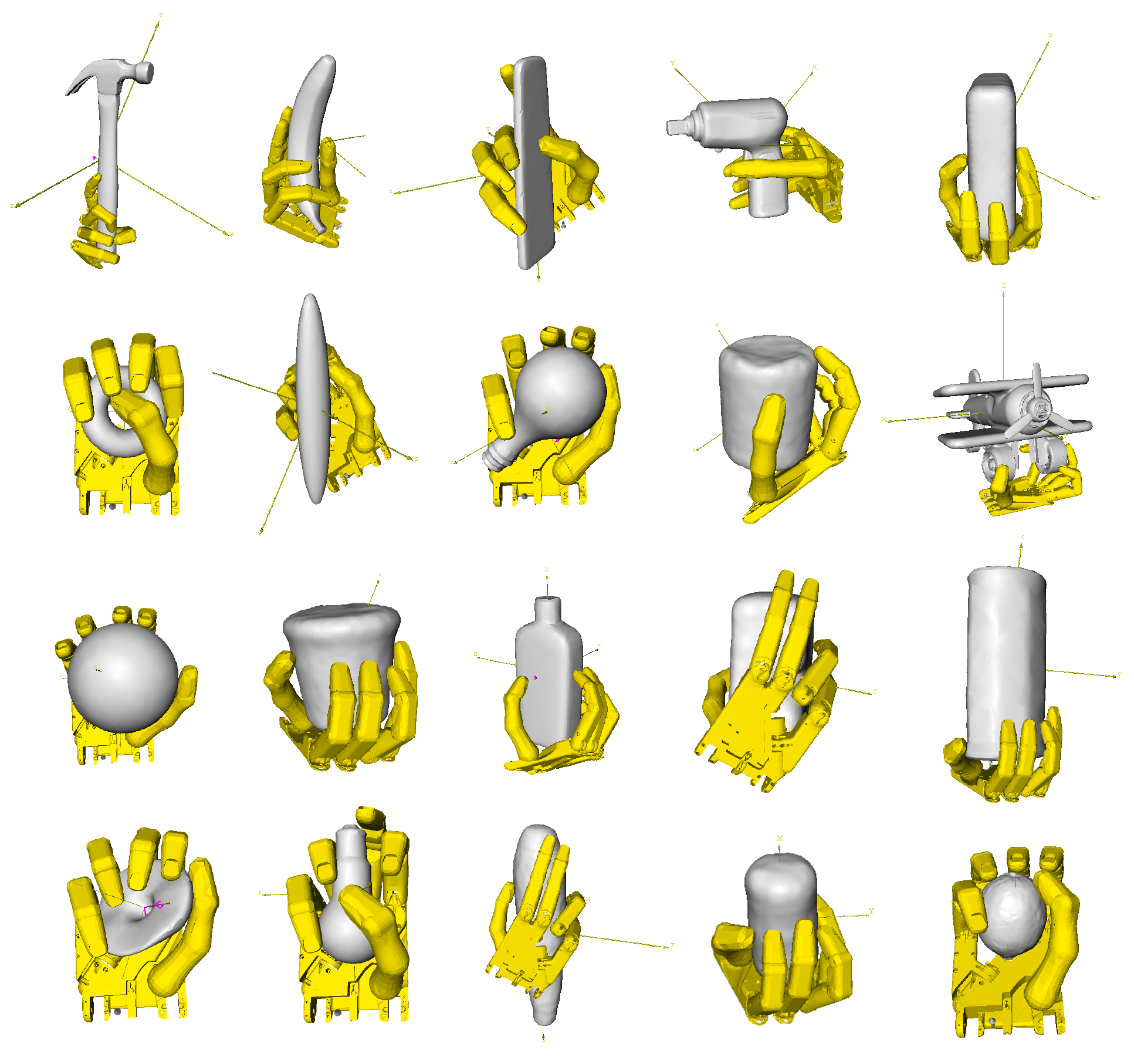}
\vspace{-10px}
\caption{Stable grasp poses generated for a large collection of unknown objects.}
\vspace{-10px}
\label{fig:gallery}
\end{figure}

The main difference between our method and prior works \cite{Fang2018MultiTaskDA,Yan2018Learning6G,8304630} is that we target high-DOF grasp poses and we use a neural network to generate grasp poses directly instead of using the score of a candidate grasp pose. Our method still needs a sampling-based algorithm to randomly rotate the target object and pick the best grasp. However, unlike \cite{Fang2018MultiTaskDA}, which requires hundreds of samples, our method only needs 3-5 samples, which can be computed within 3-5 seconds. On the other hand, a major drawback of our method is that we require a very large dataset, with tens of grasp poses for each target object. We find this dataset to be an essential component of making our method robust when generating grasp poses for unseen objects, as shown in \prettyref{fig:gallery}. Most of the generated grasp poses (after pose refinement) for unseen objects are of qualities similar to poses generated from \GRASPIT, as shown in \prettyref{fig:comparison}.

\section{\label{sec:conclusion}Conclusion and Limitations}
We present a new neural-network architecture and a training technique for the generation of high-DOF grasp poses. To resolve grasp pose redundancy, we use a consistency loss function and let the neural network pick the best or most-representable grasp poses for each target object. To further improve the quality of the grasp poses, we introduce a collision loss function to resolve penetrations between the hand and the object. Our results show that conventional supervised learning will not result in accurate grasp poses while a neural network trained using our consistency loss function drastically improves the accuracy of grasp poses compared to conventional  supervised  learning. Further, the collision loss can effectively resolve penetrations between the gripper and the target object on both the training set and the test set.

A major limitation of our current method is that it requires a very large dataset with many effective grasp poses for each target object. This is essential for the neural network to select consistent grasp poses. However, when we have a very large set of target objects, generating such a dataset can be very computationally costly and lots of computations can be wasted because the computed grasp poses are not selected by the neural network. Another limitation is that our one-to-one mapping method can cause a reachability problem. When a grasp pose is not reachable or collision-free, we have to rotate the target object grid and then feed it to our neural network until we find a feasible grasp pose.

There are several avenues for future work. One is to consider an end-to-end architecture that predicts grasp poses directly from multi-view depth images, similar to \cite{Yan2018Learning6G}. Another direction is to consider more topologically complex target objects, such as high-genus models. In these cases, a signed distance representation is not enough to resolve the geometric details of objects and the collision loss needs to be reformulated. Finally, our current method has been evaluated on a single high-DOF gripper model (the Shadow Hand) and it would be useful to generalize the ability of the neural network to represent grasp poses for other high-DOF gripper models such as a humanoid hand model, in which case we could utilize a prior method \cite{betancourt2015towards} to generate humanoid grasp data.
\section{\label{sec:ac}Acknowledgements}
We thank the anonymous reviewers for their valuable comments. This work was supported in part by NSFC (61572507, 61532003, 61622212) and Intel. Min Liu
is supported by the China Scholarship Council.
\vspace{-5px}
\bibliographystyle{IEEEtranS}
\bibliography{reference}
\end{document}